\documentclass[runningheads]{llncs}
\usepackage{setspace}
\usepackage[T1]{fontenc}
\usepackage{amssymb,amsmath}
\usepackage{graphicx}
\usepackage{graphicx}
\usepackage{algorithm}
\usepackage{algpseudocode}
\newtheorem{assumption}{Assumption}
\usepackage{multirow}
\usepackage[table,xcdraw]{xcolor}
\usepackage{hyperref}
\hypersetup{colorlinks,allcolors=blue}

\begin{document}

\title{Almost Linear Time Consistent Mode Estimation and Quick Shift Clustering}
\titlerunning{Fast Mode Estimation and Clustering}
\author{Sajjad Hashemian\inst{1}\orcidID{0009-0005-9451-5957}}
\authorrunning{S. Hashemian}

\institute{
University of Tehran, Tehran, Iran \\ \email{sajjadhashemian@ut.ac.ir}
}

\maketitle              
\begin{abstract}
In this paper, we propose a method for density-based clustering in high-dimensional spaces that combines Locality-Sensitive Hashing (LSH) with the Quick Shift algorithm. The Quick Shift algorithm, known for its hierarchical clustering capabilities, is extended by integrating approximate Kernel Density Estimation (KDE) using LSH to provide efficient density estimates. The proposed approach achieves almost linear time complexity while preserving the consistency of density-based clustering.

\keywords{Density Based Clustering  \and Locality Sensitive Hashing \and Quick Shift Clustering \and Mode Estimation}
\end{abstract}

\section{Introduction}

Density-based clustering algorithms are fundamental tools in data analysis due to their ability to identify clusters of arbitrary shapes. The most popular density-based clustering method is DBSCAN~\cite{ester1996density}\cite{schubert2017dbscan}, which defines clusters based on the concept of "density-reachability." Mean Shift~\cite{cheng1995mean}\cite{comaniciu2002mean}\cite{jang2021meanshift++} is another density-based clustering algorithm that moves each point to the densest area in its vicinity, based on kernel density estimation which is computationally challenging due to its iterative nature and the need for density estimation, making it much less scalable than DBSCAN.

To overcome this issue, Quick Shift~\cite{vedaldi2008quick}\cite{jiang2017consistency} generalizes Mean Shift by constructing a hierarchical clustering tree on density estimates. However, these methods are computationally expensive, making them less scalable for large datasets.

Jiang~\cite{jiang2017consistency} established the consistency of the Quick Shift algorithm. This result allows the consistency analysis for various density estimators, including k-nearest neighbor~\cite{dasgupta2014optimal}. Esfandiari et al.~\cite{esfandiari2021DBSCAN} demonstrated the use of certain types of LSH for fast density estimation, enabling efficient density-based clustering. Jang and Jiang~\cite{jang2019dbscan++} introduced DBSCAN++, a modification of DBSCAN that computes densities for a subset of points, reducing computational cost while maintaining performance. Xu and Pham~\cite{xu2024scalable} proposed sDBSCAN, a scalable density-based clustering algorithm in high dimensions using random projections. These works highlight the importance of efficient density estimation for scalable clustering.

Building upon these, we utilize hashing-based kernel density estimators~\cite{charikar2017hashing}\cite{charikar2020kernel} to develop a fast and consistent mode estimator and extend the Quick Shift algorithm as a result, achieving almost linear time complexity while preserving consistency.

\section{Preliminaries}

Throughout this paper \(X^{(n)} = \{x_1, x_2, \ldots, x_n\} \subset \mathbb{R}^d\) denotes a dataset of \(n\) points. We assume that the data points are drawn i.i.d.\ from a probability distribution \(F\) with density \(f\) supported on a compact set \(\mathcal{X} \subset \mathbb{R}^d\). 

\subsection{Locality Sensitive Hashing}

The Near Neighbor Search (NNS) problem is a fundamental problem in data science and computational geometry. Given a dataset \(X\), the goal is to preprocess the data such that, given a query point \(q\) in the supported set, we can efficiently return a point \(p \in X\) near to it, or report that no such point exists.

Classically known time-efficient data structures for exact NNS require space exponential in the dimension \(d\), which is prohibitively expensive for high-dimensional datasets. To address this, the \((c, r)\)-Approximate Near Neighbor Search problem (\((c, r)\)-ANN) was introduced.

\begin{definition}[\((k, c, r)\)-ANNS]
Given dataset \(X\), distance threshold \(r > 0\), and approximation factor \(c > 1\), the goal is to return \(k\) points \(p_1, p_2, \dots, p_k \in X\) such that \(d_{\mathcal{X}}(q, p_i) \leq c r\) for each \(i\), given a query point \(q\) with the promise that there are at least \(k\) points in \(X\) within distance \(r\) of \(q\).
\end{definition}

Approximate Near Neighbor Search allows for efficient data structures with query time sublinear in \(n\) and polynomial dependence on \(d\). A classic technique for solving \((c, r)\)-ANN is \textbf{Locality-Sensitive Hashing (LSH)}, introduced by Indyk and Motwani~\cite{indyk1998approximate}. The main idea behind LSH is to use random space partitions such that pairs of points within distance \(r\) are more likely to be hashed to the same bucket than pairs of points at a distance greater than \(c r\).

\begin{proposition}[Optimal LSH for \((c, r)\)-ANNS~\cite{o2014optimal}]
\label{prop: optimal-lsh}
For the Euclidean metric \(\ell_2\) and any fixed \(r > 0\), LSH yields data structures for solving \((k, c, r)\)-ANNS with space \(O(n^{1 + \rho} + d n)\) and query time \(O(d n^{\rho})\), where \(\rho = \frac{1}{c^2} - o(1)\).
\end{proposition}

\subsection{Kernel Density Estimation}

Kernel Density Estimation (KDE) is a widely used method in non-parametric statistics for estimating the probability density function of a dataset. Given a set of \(n\) points \(X \subset \mathbb{R}^d\) sampled from an unknown distribution \(F\), the goal is to estimate the density at an arbitrary point \(x \in \mathbb{R}^d\).

\begin{definition}[Kernel Density Estimation]
Given a kernel function \(k_{\sigma}: \mathbb{R}^d \times \mathbb{R}^d \to [0, 1]\) and a dataset \(X \subset \mathbb{R}^d\) of \(n\) points, the kernel density of \(X\) at a point \(x \in \mathbb{R}^d\) is defined as:
\[
K_X(x) := \frac{1}{n} \sum_{y \in X} k_{\sigma}(x, y)
\]
where \(k_{\sigma}(x, y)\) is typically a function of the Euclidean distance \(\|x - y\|\).
\end{definition}

The \textbf{Gaussian kernel} is one of the most commonly used kernels:
\[
k_{\sigma}(x, y) = \exp\left(-\frac{\|x - y\|^2}{\sigma^2}\right)
\]

The exact computation of KDE requires \(O(n^2)\) time, which is impractical for large datasets. However, approximate KDE can be computed more efficiently using techniques such as Locality-Sensitive Hashing.

\begin{proposition}[Approximate KDE via LSH~\cite{charikar2020kernel}]
\label{prop: LSH-KDE}
Given a Gaussian kernel \(k_\sigma(p, q) = \exp\left(-\frac{\|p - q\|_2^2}{\sigma^2}\right)\) for any \(\sigma > 0\), \(\epsilon = \Omega\left(\frac{1}{\operatorname{polylog} n}\right)\), \(\mu = n^{-\Theta(1)}\), and a set of points \(X\), there exists an algorithm that uses LSH to approximate \(K_X(q)\) up to a \((1 \pm \epsilon)\) multiplicative factor in time \(\widetilde{O}(\epsilon^{-2} \mu^{-o(1)})\) for any query point \(q\).
\end{proposition}

\section{Algorithm}
In this section, we present our proposed variant of QuickShift algorithm, which achieves efficient clustering with almost linear time complexity and space complexity using approximate KDE, making it suitable for large-scale high-dimensional datasets.

\begin{algorithm}
\label{alg: LSH-QuickShift}
\caption{LSH-QuickShift}

\textbf{Input:} Dataset \( X = \{x_1, x_2, \dots, x_n\} \subset \mathbb{R}^d \), bandwidth parameter \( h \)
\\
\textbf{Output:} Directed graph \( G \) representing the clustering structure

\begin{algorithmic}[1]
\State \textbf{Initialize:} Directed graph \( G \) with vertices \( \{x_1, x_2, \dots, x_n\} \) and no edges
\State \textbf{Initialize:} Preprocess $X$ for $(c, h)$-ANNS query.
\For{each point \( x_i \in X \)}
    \State Compute the approximate KDE \( \tilde{f}(x_i) \)
\EndFor
\For{each point \( x_i \in X \)}
    \State \(\hat{x_i}=\arg\max_{x_j\in (c,h)-ANNS(x_i)}\tilde{f}(x_i)\)
    \If{ \( \tilde{f}(\hat{x_i}) > \tilde{f}(x_i) \) }
        \State Add a directed edge from \( x_i \) to \( \hat{x_i} \) in \( G \)
    \EndIf
\EndFor
\State \textbf{Return} Directed graph \( G \)
\end{algorithmic}
\end{algorithm}

\begin{theorem}[Computational Complexity]
Providing a dataset $X\subset \mathbb{R}^d$,
there exist an algorithm (Algorithm~\ref{alg: LSH-QuickShift}) that performs density based clustering in time and space $O(dn^{1+o(1)})$.
\end{theorem}
\begin{proof}
The time and space complexity of the LSH-QuickShift algorithm is determined by LSH Preprocessing, KDE, and graph construction steps.

Constructing the hash tables for Locality-Sensitive Hashing requires \( O(n^{1+\rho} + dn) \) time and space, where \( n \) is the number of data points, \( d \) is the dimensionality, and \( \rho = \frac{1}{c^2} - o(1) \) for the given approximation factor using the optimal LSH using proposition~\ref{prop: optimal-lsh}.

For each point \( x_i \), the approximate kernel density estimate \( \tilde{f}(x_i) \) is computed using the LSH-KDE oracle, with the total time of \( \tilde{O}(n) \) as proposition ~\ref{prop: LSH-KDE}.

Finally, the algorithm iterates over all $x_i\in X$ and add a directed edge to a point with higher estimated density in $G$. We can do this efficiently by keeping the maximum over all non-empty hash key, then we can answer this type of query with the same complexity as the $(c,h)-$-ANNS.

Combining these components and assuming that $c\leq \tau_m$ is a constant (Assumption~\ref{assump:modes}), the overall time complexity is $O(dn^{1+o(1)})$. Also, the space complexity is dominated by the LSH preprocessing step which provides the same bound and completes the proof.
\qed
\end{proof}

\section{Theoretical Analysis}
In this section we show that, despite the multiplicative error introduced by the approximate KDE in proposition~\ref{prop: LSH-KDE}, Quick Shift’s assignment of points to mode-rooted trees remains consistent with the underlying density.

These properties form the foundation for consistent clustering guarantees, analogous to those established in the exact KDE setting~\cite{jiang2017consistency}, but now achieved with sub-quadratic computational complexity by virtue of using the approximate KDE as an oracle.

\begin{assumption}[Hölder Density]\label{assump:holder}
There exist constants \(0<\alpha\le 1\) and \(C_\alpha>0\) such that for all \(x,x'\in \mathcal{X}\),
\[
|f(x)-f(x')|\le C_\alpha \|x-x'\|^\alpha.
\]
\end{assumption}

\begin{assumption}[Kernel Properties]\label{assump:kernel}
Let \(K:\mathbb{R}^d\to \mathbb{R}_{\ge 0}\) be a kernel function such that there exists a non-increasing function \(k:[0,\infty)\to\mathbb{R}_{\ge 0}\) with 
\(
K(u)=k(\|u\|)
\)
such that,
\[
\int_{\mathbb{R}^d}K(u)\,du=1.
\]
and assume that there exist constants \(\rho, C_\rho, t_0>0\) such that for all \(t>t_0\),
\[
k(t)\le C_\rho\exp(-t^\rho).
\]
\end{assumption}

For a given bandwidth \(h>0\), the classical kernel density estimator (KDE) is defined by
\[
\hat{f}_h(x)=\frac{1}{n\,h^d}\sum_{i=1}^n K\!\Bigl(\frac{x-x_i}{h}\Bigr).
\]

\begin{assumption}[LSH-KDE Oracle]
\label{assump:lsh}
An LSH-based KDE oracle \(O_{\mathrm{LSH\mbox{-}KDE}}\) is available that, for any query \(q\in \mathbb{R}^d\) and any prescribed error \(\epsilon>0\), returns an approximation \(\tilde{f}(q)\) satisfying
\[
(1-\epsilon)\,\hat{f}_h(q)\le \tilde{f}(q)\le (1+\epsilon)\,\hat{f}_h(q),
\]
with probability at least \(1-1/n\) for all \(q\in\mathcal{X}\), provided that \(h\ge (\log n/n)^{1/d}\).
\end{assumption}

Under Assumptions~\ref{assump:holder}--\ref{assump:lsh}, standard uniform convergence arguments (see, e.g.,~\cite{jiang2017consistency}) implies that there exists a constant \(C'>0\) such that with probability at least \(1-1/n\)
\[
\sup_{x\in\mathcal{X}} \Bigl|\hat{f}_h(x)-f(x)\Bigr| \le C'\Bigl( h^\alpha + \sqrt{\frac{\log n}{n\,h^d}} \Bigr).
\]
By the accuracy guarantee of the oracle, for sufficiently small \(\epsilon>0\) there exists a constant \(C''>0\) so that
\[
\sup_{x\in\mathcal{X}} \bigl|\tilde{f}(x)-f(x)\bigr| \le \delta_n, \quad \text{with} \quad \delta_n = C''\left( h^\alpha + \sqrt{\frac{\log n}{n\,h^d}} \right).
\]

\subsection{Mode Estimation}

We now describe the mode estimation problem under the Quick Shift clustering procedure, where density evaluations are performed using the LSH-KDE oracles.

Let \(M\subset \mathcal{X}\) denote the set of local modes of \(f\). We assume that modes are isolated and exhibit quadratic decay.

\begin{assumption}[Modes]\label{assump:modes}
A point \(x_0\in \mathcal{X}\) is said to be a mode of \(f\) if there exists \(r_M>0\) such that \(x_0\) is the unique maximizer of \(f\) in \(B(x_0,r_M)\) and there exist constants \(\check{C},\hat{C}>0\) for which
\[
\check{C}\,\|x-x_0\|^2 \le f(x_0)-f(x) \le \hat{C}\,\|x-x_0\|^2, \quad \forall\, x\in B(x_0,r_M).
\]
Denote by \(M\) the (finite) set of all such modes.
\end{assumption}

Quick Shift (see, e.g.,~\cite{vedaldi2008quick}) is an iterative procedure that assigns each sample \(x_i\) to a nearby point in its \(\tau\)-ball with strictly higher density. In our algorithm, each density evaluation is computed via \(\tilde{f}(x)\). We denote by \(\hat{M}\) the set of estimated modes returned by the algorithm.

\begin{theorem}[Mode estimation via LSH-KDE Quick Shift]\label{thm:mode}
Let Assumptions~\ref{assump:holder}, ~\ref{assump:kernel}, ~\ref{assump:lsh} and~\ref{assump:modes} hold. Suppose that the bandwidth \(h=h(n)\) satisfy
\[
h\to 0 \quad \text{and} \quad \frac{\log n}{n\,h^d}\to 0 \quad \text{as } n\to\infty.
\]
Then there exists a constant \(C>0\) such that with probability at least \(1-1/n\) the Hausdorff distance between the true mode set \(M\) and the estimated mode set \(\hat{M}\) satisfies
\[
d_H(M,\hat{M})^2 \le C\left( \frac{(\log n)^4}{h^2} + \sqrt{\frac{\log n}{n\,h^d}} \right).
\]
\end{theorem}

\begin{proof}
Let
\(
\hat{f}_h(x)=\frac{1}{n\,h^d}\sum_{i=1}^n K\!\Bigl(\frac{x-x_i}{h}\Bigr)
\)
be the classical KDE. Under Assumptions~\ref{assump:holder} and~\ref{assump:kernel}, standard results (e.g., Theorem 1,~\cite{jiang2017consistency}) guarantee that with probability at least \(1-1/n\),
\[
\sup_{x\in\mathcal{X}} \bigl|\hat{f}_h(x)-f(x)\bigr| \le C'\left( h^\alpha + \sqrt{\frac{\log n}{n\,h^d}} \right)
\]
for some constant \(C'>0\). By Assumption~\ref{assump:lsh}, the LSH-KDE oracle returns an approximation \(\tilde{f}(x)\) satisfying
\(
(1-\epsilon)\,\hat{f}_h(x)\le \tilde{f}(x)\le (1+\epsilon)\,\hat{f}_h(x)
\)
with high probability. Hence, for sufficiently small \(\epsilon>0\), there exists \(C''>0\) such that
\[
\sup_{x\in\mathcal{X}} \bigl|\tilde{f}(x)-f(x)\bigr| \le \delta_n, \quad \text{with} \quad \delta_n = C''\left( h^\alpha + \sqrt{\frac{\log n}{n\,h^d}} \right).
\]

Let \(x_0\in M\) be a true mode. By Assumption~\ref{assump:modes}, there exists \(r_M>0\) and constants \(\check{C},\hat{C}>0\) such that
\[
\check{C}\,\|x-x_0\|^2 \le f(x_0)-f(x) \le \hat{C}\,\|x-x_0\|^2, \quad \forall\, x\in B(x_0,r_M).
\]
Since \(\tau < r_M/2\), the ball \(B(x_0,\tau)\) is contained in \(B(x_0,r_M)\), define
\[
\hat{x} = \arg\max_{x\in B(x_0,\tau)} \tilde{f}(x).
\]
Then,
\(
\tilde{f}(x_0) \ge f(x_0)-\delta_n.
\)
And, for any \(x\in B(x_0,\tau)\) with \(\|x-x_0\|\ge \eta\) (for some \(\eta>0\) to be determined), the quadratic decay of \(f\) yields
\[
f(x)\le f(x_0)-\check{C}\,\|x-x_0\|^2 \le f(x_0)-\check{C}\,\eta^2,
\]
\[
\tilde{f}(x) \le f(x) + \delta_n \le f(x_0)-\check{C}\,\eta^2+\delta_n.
\]
Now, if we require that
\(
\check{C}\,\eta^2>2\delta_n,
\)
then
\[
\tilde{f}(x) \le f(x_0)-\check{C}\,\eta^2+\delta_n < f(x_0)-\delta_n \le \tilde{f}(x_0).
\]
Since \(\hat{x}\) maximizes \(\tilde{f}\) on \(B(x_0,\tau)\), it follows that
\(
\|\hat{x}-x_0\|<\eta.
\)
Thus, by choosing \(\eta = \sqrt{2\delta_n/\check{C}}\) implies
\(
\|\hat{x}-x_0\| \le \sqrt{\frac{2\delta_n}{\check{C}}}.
\)
In particular, if we set \(\alpha=1\) (or if \(h^\alpha\) and the stochastic term are of the same order) and choose \(h\asymp n^{-1/(4+d)}\), then
\[
\|\hat{x}-x_0\| = \widetilde{O}\Bigl(n^{-1/(4+d)}\Bigr).
\]
A symmetric argument shows that every \(\hat{x}\in\hat{M}\) is associated uniquely to a true mode \(x_0\in M\). Consequently, the Hausdorff distance satisfies
\[
d_H(M,\hat{M})^2 \le \frac{2C''}{\check{C}}\left( h^\alpha + \sqrt{\frac{\log n}{n\,h^d}} \right).
\]
A refinement of this argument, via a union bound over the sample yields the stated bound
\[
d_H(M,\hat{M})^2 \le C\left( \frac{(\log n)^4}{h^2} + \sqrt{\frac{\log n}{n\,h^d}} \right),
\]
for some constant \(C>0\) which completes the proof.\qed
\end{proof}

\subsection{Assignment of Points to Modes}
In this section, we show how Quick Shift assigns each sample point to the \textit{basin of attraction} of a nearby mode under the approximate KDE oracle. In particular, we establish that if two points are separated by a sufficiently deep and wide valley in the underlying density, they cannot lie in the same directed tree of the Quick Shift graph.

\begin{definition}[$(r,\delta)$-separation,~\cite{dasgupta2014optimal}]
\label{def:r-delta-sep}
Let $r>0$ and $\delta>0$. Two points $x_1, x_2 \in \mathcal{X}$ are said to be \emph{$(r,\delta)$-separated} if there exists a set $S \subset \mathcal{X}$ such that:
\begin{enumerate}
\item Every path from $x_1$ to $x_2$ intersects $S$.
\item We have
\[
\sup_{x \in S + B(0,r)} f(x)
\;\;<\;\;
\inf_{x\in B(x_1,r)\,\cup\, B(x_2,r)} f(x) \;-\;\delta.
\]
\end{enumerate}
\end{definition}

Equivalently, one may view $S$ as the region forming a ``deep valley'' separating $x_1$ from $x_2$, whose density is at least $\delta$ below the density near $x_1$ or $x_2$, plus a margin of width $r$.
The main technical claim is that if two points $x_1$ and $x_2$ are $(r_s,\delta)$-separated, then with high probability there is no directed path in $(G,\tilde{f})$ from $x_1$ to $x_2$. 

\begin{theorem}\label{thm: sepration}
Let $(G,\tilde{f})$ be the directed cluster tree constructed by Quick Shift using the approximate densities $\tilde{f}$ from Assumption~\ref{assump:lsh}. Under the same conditions as Theorem~\ref{thm:mode}, there exists a constant $C>0$ such that with probability at least $1-1/n$, if $x_1$ and $x_2$ are $(r_s,\delta)$-separated, there is no directed path from $x_1$ to $x_2$ in $G$, provided $\delta > C\,\epsilon \,\sup_{x\in\mathcal{X}} \hat{f}(x)$.
\end{theorem}

\begin{proof}
Suppose, for contradiction, that there is a directed path $x_1 = y_0 \to y_1 \to \cdots \to y_k = x_2$ in $G$. By definition of Quick Shift, each edge $(y_j, y_{j+1})$ implies
\[
\tilde{f}(y_{j+1}) \;>\; \tilde{f}(y_j)
\quad\text{and}\quad
\|y_{j+1}-y_j\|\le \tau.
\]
Since $\tau < r_s/2$, the entire path is made of steps of radius at most $\tau$.

By Definition~\ref{def:r-delta-sep}, every path from $x_1$ to $x_2$ intersects the seperator set $S + B(0,r_s)$. Hence, there exists at least one point $y_j^*\in S + B(0,r_s)$ on the path. By $(r_s,\delta)$-separation,
\[
f(y_j^*) 
\;\;\le\;\;
\sup_{x \,\in\, S + B(0,r_s)} f(x)
\;\;<\;\;
\inf_{x\in B(x_1,r_s)\cup B(x_2,r_s)} f(x) \;-\;\delta.
\]
Thus, if $\|x_1 - y_j^*\| \le \|x_1 - x_2\| + \|x_2 - y_j^*\|$, then in fact such a $y_j^*$ is forced to be on any path that attempts to connect $x_1$ to $x_2$. Now, if we compare $\tilde{f}(y_j^*)$ and $\tilde{f}(x_1)$, by KDE uniform bounds~\cite{jiang2017uniform},
\[
\hat{f}_h(y_j^*) \;\le\; f(y_j^*) + O\!\Bigl(h^\alpha + \sqrt{\tfrac{\log n}{n\,h^d}}\Bigr)
\;\;\le\;\;
f(x_1) - \frac{\delta}{2},
\]
So if $n$ is sufficiently large, provided that $\delta$ is chosen larger than a constant times the error term, applying the $(1\pm \epsilon)$ approximation in Assumption~\ref{assump:lsh}, we obtain
\[
\tilde{f}(y_j^*)
\;\le\;
(1+\epsilon)\,\hat{f}_h(y_j^*)
\;\le\;
(1+\epsilon)\,\Bigl(f(x_1) - \tfrac{\delta}{2}\Bigr).
\]
On the other hand,
\[
\tilde{f}(x_1)
\;\ge\;
(1-\epsilon)\,\hat{f}_h(x_1)
\;\approx\;
(1-\epsilon)\,f(x_1),
\]
and for $\delta$ sufficiently large relative to $\epsilon\, f(x_1)$, we deduce
\(
\tilde{f}(y_j^*) \;<\; \tilde{f}(x_1).
\),
Which contradicts the requirement for a directed edge path that $\tilde{f}(y_{j+1}) > \tilde{f}(y_j)$ strictly at each step. Hence, no such path can exist from $x_1$ to $x_2$.

\qed
\end{proof}

An immediate corollary is that points $(r_s,\delta)$-separated from a given mode $x_0$ cannot join $x_0$'s tree in $G$. Indeed, if $x$ were assigned to $x_0$, there would exist a directed path $x \to \cdots \to x_0$ in $(G,\tilde{f})$, violating Theorem~\ref{thm: sepration}. Hence, each point is constrained to remain in the basin of attraction of exactly those modes \emph{not} separated from it by a deep valley.

\begin{corollary}[Separation Implies Different Trees]
\label{cor:sep-trees}
Under the same conditions as Theorem~\ref{thm: sepration}, if $x$ and a mode $x_0$ are $(r_s,\delta)$-separated, then $x$ cannot lie in the Quick Shift tree rooted at $x_0$.
\end{corollary}

\begin{proof}
If $x$ were in the tree of $x_0$, then there would be a directed path $x \to \cdots \to x_0$. This contradicts Theorem~\ref{thm: sepration} because $x$ and $x_0$ are $(r_s,\delta)$-separated, implying no such path can exist. \qed
\end{proof}

In the absence of approximation error (i.e.\ $\epsilon=0$), these results coincide exactly with the analysis in~\cite{jiang2017consistency}. The only additional requirement here is that $\delta$ exceed a constant times $\epsilon$ (and the usual KDE deviation term), ensuring that approximate comparisons preserve the strict density gap. Thus, the \emph{same} geometric intuition that ``deep and wide valleys'' prevent two points from being assigned to the same root remains valid under approximate density evaluations.

\section{Experiments}
We evaluate the proposed algorithm on two tasks: clustering, and image segmentation on benchmark data as the most well-known tasks for Mean Shift variants. All experiments were conducted on a standard workstation (16 GB RAM, 2.4 GHz CPU).
\subsection{Clustering}
We compare the proposed algorithm against other popular density-based clustering algorithm and K-means as an scalable option on various clustering tasks in Table~\ref{tabclustering}. These comparisons are made using the Scikit-Learn\footnote{\href{https://scikit-learn.org}{\color{blue}Scikit-Learn Webpage}} and Scikit-Image \footnote{\href{https://scikit-image.org}{\color{blue}Scikit-Image Webpage}} implementation of Quick Shift and our own implementation of LSH-Quickshift in C++ wrapped as a Python package using Pybind11, and used FAISS library~\cite{faiss} for LSH implemention. To measure the quality of a clustering result, we use the Adjusted Rand Index (ARI)~\cite{ARI} and the Adjusted Mutual Information (AMI)~\cite{AMI} scores, comparing the clustering with the partitioning induced by the labels of the data points. The benchmark datasets we use are labeled datasets from UCI Repository \footnote{\href{https://archive.ics.uci.edu}{\color{blue}UCI Machine Learning Repository Webpage}},
and we only cluster the features.

As expected by our theoretical analysis, the proposed method is well scalable for large data sets in high dimensions. Also, this method outpreforms vector quantization methods in complex regimes with high number of clusters. 

\subsection{Image segmentation}
\begin{figure}[ht!]
    \hspace{-20pt}
    \includegraphics[width=1.1\linewidth]{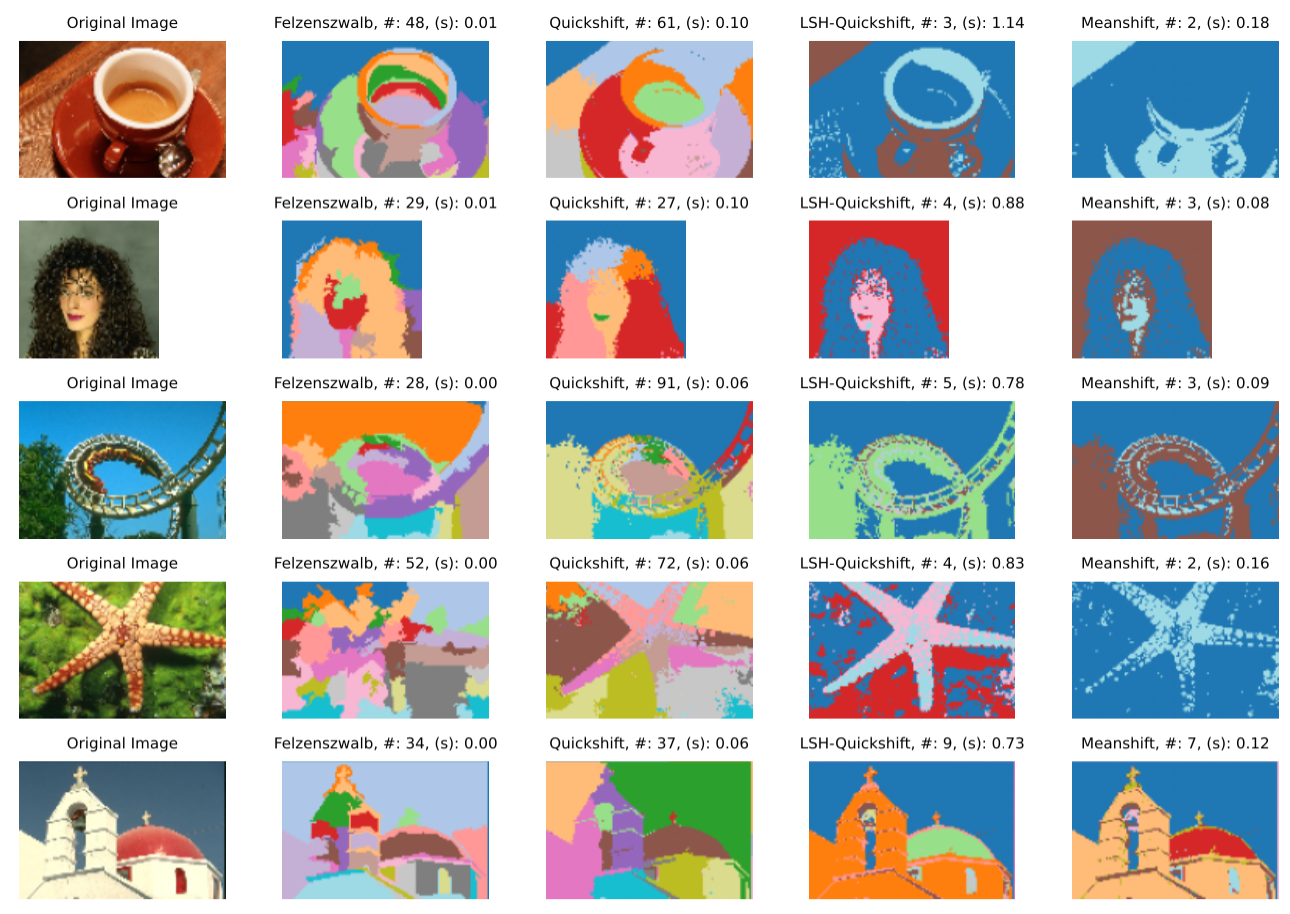}
    \caption{Comparison of image segmentation algorithms. For each image, the number of detected segments (\#), computation time (in seconds), and the segmentation method are indicated above.}
    \label{fig:segmentation}
\end{figure}

We compare the proposed method to a number of baselines for unsupervised image segmentation in Figure~\ref{fig:segmentation}. We include Felzenszwalb~\cite{felzenszwalb2004efficient}, Quick Shift~\cite{vedaldi2008quick}, and Mean Shift~\cite{cheng1995mean}, three popular image segmentation procedures from the Python, Scikit-Image library.
For image segmentation, we run each algorithm on a preprocessed image with each pixel represented in a 5D $(r,g,b,x,y)$ color channel and spatial coordinates space and at maximum the size of our images was 46,500 pixels from the Berkeley Segmentation Dataset Benchmark (BSDS500) \footnote{\href{https://www2.eecs.berkeley.edu/Research/Projects/CS/vision/bsds/}{\color{blue}Berkeley Segmentation Dataset Webpage}}. For each algorithm, the returned clusters are taken as the segments.
Our image segmentation experiments show that LSH-Quick Shift is able to produce segmentations that are nearly identical to that of Mean Shift.

\begin{table}[!ht]
\centering
\begin{tabular}{|c|c|c|c|c|c|c|c|c|}
\cline{1-9}
Dataset                                                       & n                       & d                     & C                    & DBSCAN                                 & KMeans                                   & LSH-QS  & MeanShift                              & QuickShift  \\ 
\cline{1-9}
{\color[HTML]{0000EE} }                                       &                         &                       &                      & 0.047                                  & {\color[HTML]{006300} \textbf{0.1196}}   & {\color[HTML]{C68000} \textbf{0.113}}   & 0.0442                                 & 0.0266                                       \\
{\color[HTML]{0000EE} }                                       &                         &                       &                      & 0.0486                                 & {\color[HTML]{C68000} \textbf{0.4991}}   & {\color[HTML]{006300} \textbf{0.1993}}  & 0.0698                                 & 0.0556                                      \\

\multirow{-3}{*}{\href{https://archive.ics.uci.edu/dataset/254/qsar+biodegradation}{\color{blue}biodegradation}} & \multirow{-3}{*}{1054}  & \multirow{-3}{*}{41}  & \multirow{-3}{*}{2}  & {\color[HTML]{006300} \textbf{0.0627}} & {\color[HTML]{C68000} \textbf{0.0731}}   & 0.1631                                  & 31.8839                                & 0.2399                                  \\ \cline{1-9}
{\color[HTML]{0000EE} }                                       &                         &                       &                      & 0.0039                                 & {\color[HTML]{006300} \textbf{0.6703}}   & {\color[HTML]{C68000} \textbf{0.5361}}  & 0.0079                                 & 0.0005                                       \\
{\color[HTML]{0000EE} }                                       &                         &                       &                      & 0.0001                                 & {\color[HTML]{006300} \textbf{0.5363}}   & {\color[HTML]{C68000} \textbf{0.3719}}  & 0.0001                                 & 0                                           \\

\multirow{-3}{*}{\href{https://archive.ics.uci.edu/ml/datasets/Optical+Recognition+of+Handwritten+Digits}{\color{blue}digits}}         & \multirow{-3}{*}{1797}  & \multirow{-3}{*}{64}  & \multirow{-3}{*}{10} & {\color[HTML]{006300} \textbf{0.0823}} & {\color[HTML]{C68000} \textbf{0.1266}}   & 0.4561                                  & 47.9048                                & 0.4432                                  \\ \cline{1-9}
{\color[HTML]{0000EE} }                                       &                         &                       &                      & 0.1                                    & {\color[HTML]{006300} \textbf{0.6277}}   & {\color[HTML]{C68000} \textbf{0.4025}}  & 0.1039                                 & 0.1039                                       \\
{\color[HTML]{0000EE} }                                       &                         &                       &                      & 0.0387                                 & {\color[HTML]{006300} \textbf{0.5179}}   & {\color[HTML]{C68000} \textbf{0.374}}   & 0.0381                                 & 0.0381                                      \\

\multirow{-3}{*}{\href{https://archive.ics.uci.edu/dataset/39/ecoli}{\color{blue}ecoli}}          & \multirow{-3}{*}{336}   & \multirow{-3}{*}{7}   & \multirow{-3}{*}{8}  & {\color[HTML]{006300} \textbf{0.0146}} & 0.076                                    & {\color[HTML]{C68000} \textbf{0.0306}}  & 1.1678                                 & 0.0413                                  \\ \cline{1-9}
{\color[HTML]{0000EE} }                                       &                         &                       &                      & {\color[HTML]{006300} \textbf{0.3626}} & {\color[HTML]{C68000} \textbf{0.1231}}   & 0.0337                                  & 0.1148                                 & 0.1171                                       \\
{\color[HTML]{0000EE} }                                       &                         &                       &                      & {\color[HTML]{006300} \textbf{0.3511}} & 0.1679                                   & 0.2653                                  & {\color[HTML]{C68000} \textbf{0.2944}} & 0.2926                                      \\

\multirow{-3}{*}{\href{https://archive.ics.uci.edu/dataset/52/ionosphere}{\color{blue}ionosphere}}     & \multirow{-3}{*}{351}   & \multirow{-3}{*}{34}  & \multirow{-3}{*}{2}  & {\color[HTML]{006300} \textbf{0.0279}} & 0.0753                                   & 0.0737                                  & 13.2798                                & {\color[HTML]{C68000} \textbf{0.0556}}  \\ \cline{1-9}
{\color[HTML]{0000EE} }                                       &                         &                       &                      & {\color[HTML]{006300} \textbf{0.7316}} & 0.6552                                   & 0.6733                                  & {\color[HTML]{006300} \textbf{0.7316}} & {\color[HTML]{006300} \textbf{0.7316}}       \\
{\color[HTML]{0000EE} }                                       &                         &                       &                      & 0.5681                                 & {\color[HTML]{C68000} \textbf{0.6201}}   & {\color[HTML]{006300} \textbf{0.6422}}  & 0.5681                                 & 0.5681                                      \\

\multirow{-3}{*}{\href{https://archive.ics.uci.edu/dataset/53/iris}{\color{blue}iris}}           & \multirow{-3}{*}{150}   & \multirow{-3}{*}{4}   & \multirow{-3}{*}{3}  & {\color[HTML]{006300} \textbf{0.0046}} & 0.0765                                   & 0.0174                                  & 0.3171                                 & {\color[HTML]{C68000} \textbf{0.011}}   \\ \cline{1-9}
{\color[HTML]{0000EE} }                                       &                         &                       &                      & NaN                                    & {\color[HTML]{006300} \textbf{0.4304}}   & {\color[HTML]{C68000} \textbf{0.1372}}  & NaN                                    & NaN                                          \\
{\color[HTML]{0000EE} }                                       &                         &                       &                      & NaN                                    & {\color[HTML]{006300} \textbf{0.3215}}   & {\color[HTML]{C68000} \textbf{0.0682}}  & NaN                                    & NaN                                         \\

\multirow{-3}{*}{\href{https://archive.ics.uci.edu/dataset/683/mnist+database+of+handwritten+digits}{\color{blue}mnist}}          & \multirow{-3}{*}{70000} & \multirow{-3}{*}{784} & \multirow{-3}{*}{10} & $\infty$                                    & {\color[HTML]{006300} \textbf{63.0059}}  & {\color[HTML]{C68000} \textbf{63.2678}} & $\infty$                                    & $\infty$                                     \\ \cline{1-9}
{\color[HTML]{0000EE} }                                       &                         &                       &                      & NaN                                    & {\color[HTML]{C68000} \textbf{0.0028}}   & {\color[HTML]{006300} \textbf{0.0036}}  & NaN                                    & NaN                                          \\
{\color[HTML]{0000EE} }                                       &                         &                       &                      & NaN                                    & {\color[HTML]{C68000} \textbf{0.0022}}   & {\color[HTML]{006300} \textbf{0.0031}}  & NaN                                    & NaN                                         \\

\multirow{-3}{*}{\href{https://github.com/zalandoresearch/fashion-mnist}{\color{blue}fashion}}        & \multirow{-3}{*}{60000} & \multirow{-3}{*}{784} & \multirow{-3}{*}{66} & $\infty$                                    & {\color[HTML]{C68000} \textbf{162.5004}} & {\color[HTML]{006300} \textbf{51.3207}} & $\infty$                                    & $\infty$                                     \\ \cline{1-9}
{\color[HTML]{0000EE} }                                       &                         &                       &                      & 0.0086                                 & {\color[HTML]{006300} \textbf{0.112}}    & {\color[HTML]{C68000} \textbf{0.0942}}  & 0.0053                                 & 0.0072                                       \\
{\color[HTML]{0000EE} }                                       &                         &                       &                      & 0.0006                                 & {\color[HTML]{C68000} \textbf{0.0769}}   & {\color[HTML]{006300} \textbf{0.0772}}  & 0.0006                                 & 0.0006                                      \\

\multirow{-3}{*}{\href{https://archive.ics.uci.edu/dataset/149/statlog+vehicle+silhouettes}{\color{blue}vehicle}}        & \multirow{-3}{*}{846}   & \multirow{-3}{*}{18}  & \multirow{-3}{*}{4}  & {\color[HTML]{006300} \textbf{0.0698}} & 0.1083                                   & {\color[HTML]{C68000} \textbf{0.0942}}  & 66.1409                                & 0.1782                                  \\ \cline{1-9}
\end{tabular}
\vspace{5pt}
\caption{
Scores of algorithms on real world benchmark datasets. Reference is provided through clickable links for each dataset.  The first row corresponds to Adjusted Mutual Information (AMI), the second row corresponds to Adjusted Rand Index, and the last row denots the computation time in seconds. Each procedure was tuned in its respective essential hyperparameter.
\\
In each row, the highest score is shown in {\color[HTML]{006300} \textbf{green}} and the second highest score in {\color[HTML]{C68000} \textbf{orange}}. As we
can see that our algorithm has the top-2 score on 17 metrics.
}
\label{tabclustering}
\vspace{-30pt}
\end{table}

\section{Conclusion}

In this work, we have introduced the LSH-QuickShift algorithm, which integrates Locality-Sensitive Hashing (LSH) with the Quick Shift clustering method to achieve efficient, provably consistent mode estimation in high-dimensional spaces. By leveraging LSH for approximate kernel density estimation, the algorithm significantly reduces computational complexity, making it suitable for large-scale datasets. Theoretical analyses confirm that the LSH-QuickShift algorithm maintains consistency with the underlying density, ensuring reliable clustering results.

However, several things remain for further consideration and applying the LSH enhanced density based clustering algorithm. Investigating methods to reconstruct the hierarchical structure of clusters post-clustering could provide deeper insights into the data's inherent organization and thus more efficient clustering algorithm. This extends the algorithm to perform regression tasks by modeling the relationship between data points and their modes and thus could broaden its applicability.

One may combine LSH-QuickShift with sub-sampling techniques and random projections and further enhance scalability and robustness, particularly in extremely high-dimensional settings with large data sets, and exploring the algorithm's performance within massive parallel computing frameworks could lead to significant improvements in processing large-scale datasets, addressing both time and space complexity challenges. 
Also, applying these methods to various data modalities beyond numerical and image data, such as text or time-series data, could demonstrate its versatility and effectiveness across different domains.

\begin{credits}
\subsubsection{\discintname}
The authors have no competing interests to declare that are relevant to the content of this article.
\end{credits}

%
\bibliographystyle{splncs04}
\bibliography{mybibliography}

\end{document}